\pdfoutput=1
\documentclass[11pt]{article}

\usepackage{emnlp2021}

\usepackage{times}
\usepackage{latexsym}
\usepackage{microtype}
\usepackage{booktabs}
\usepackage{multirow}
\usepackage{amsmath}
\usepackage{amssymb}
\usepackage{array}
\usepackage{xspace}
\usepackage{latexsym}
\usepackage{graphicx}
\usepackage{dblfloatfix}
\usepackage[linesnumbered,ruled,vlined]{algorithm2e}
\usepackage[T1]{fontenc}
\usepackage[utf8]{inputenc}

\SetKwInOut{Input}{Input}

\newcommand{\silveramr}{{\small\textsc{SilverAMR}}\xspace}
\newcommand{\silversent}{{\small\textsc{SilverSent}}\xspace}
\newcommand{\goldamr}{{\small\textsc{GoldAMR}}\xspace}

\newcommand\ftnote[1]{\footnote{\raggedright#1}}

\usepackage{todonotes}

\usepackage{microtype}

\title{Smelting Gold and Silver for Improved Multilingual \\ AMR-to-Text Generation}

\author{Leonardo F. R. Ribeiro$^{\dag}$, Jonas Pfeiffer$^{\dag}$, Yue Zhang$^{\ddag}$ and Iryna Gurevych$^{\dag}$ \vspace{1mm} \\
\rule{0pt}{2.5ex}
  $^{\dag}$Ubiquitous Knowledge Processing Lab, Technical University of Darmstadt\\
  $^{\ddag}$School of Engineering, Westlake University \\
 \texttt{ribeiro@aiphes.tu-darmstadt.de}
}

\begin{document}
\maketitle
\begin{abstract}

Recent work on multilingual AMR-to-text generation has exclusively focused on data augmentation strategies that utilize silver AMR. However, this assumes a high quality of generated AMRs, potentially limiting the transferability to the target task. In this paper, we investigate different techniques for automatically generating AMR annotations, where we aim to study which source of information yields better multilingual results. Our models trained on gold AMR with silver (machine translated) sentences outperform approaches which leverage generated silver AMR. We find that combining both complementary sources of information further improves multilingual AMR-to-text generation. Our models surpass the previous state of the art for German, Italian, Spanish, and Chinese by a large margin.\ftnote{Our code and checkpoints are available at \href{https://github.com/UKPLab/m-AMR2Text}{https://github.com/UKPLab/m-AMR2Text}.}

\end{abstract}

\section{Introduction}

AMR-to-text generation is the task of recovering a text with the same meaning as a given Abstract Meaning Representation (AMR)~\cite{banarescu-etal-2013-abstract}, and  has recently received much research interest \cite{ribeiro-etal-2019-enhancing,wang-etal-2020-amr, mager2020gpttoo, harkous2020text, fu-etal-2021-end}. AMR has applications to a range of NLP tasks, including summarization \cite{hardy-vlachos-2018-guided} and spoken language understanding~\cite{damonte-etal-2019-practical}, and has the potential power of acting as an \emph{interlingua} that allows the generation of text in many different languages~\cite{damonte-cohen-2018-cross, zhu-etal-2019-towards}.

\begin{figure}[t]
    \centering
    \includegraphics[width=.27\textwidth]{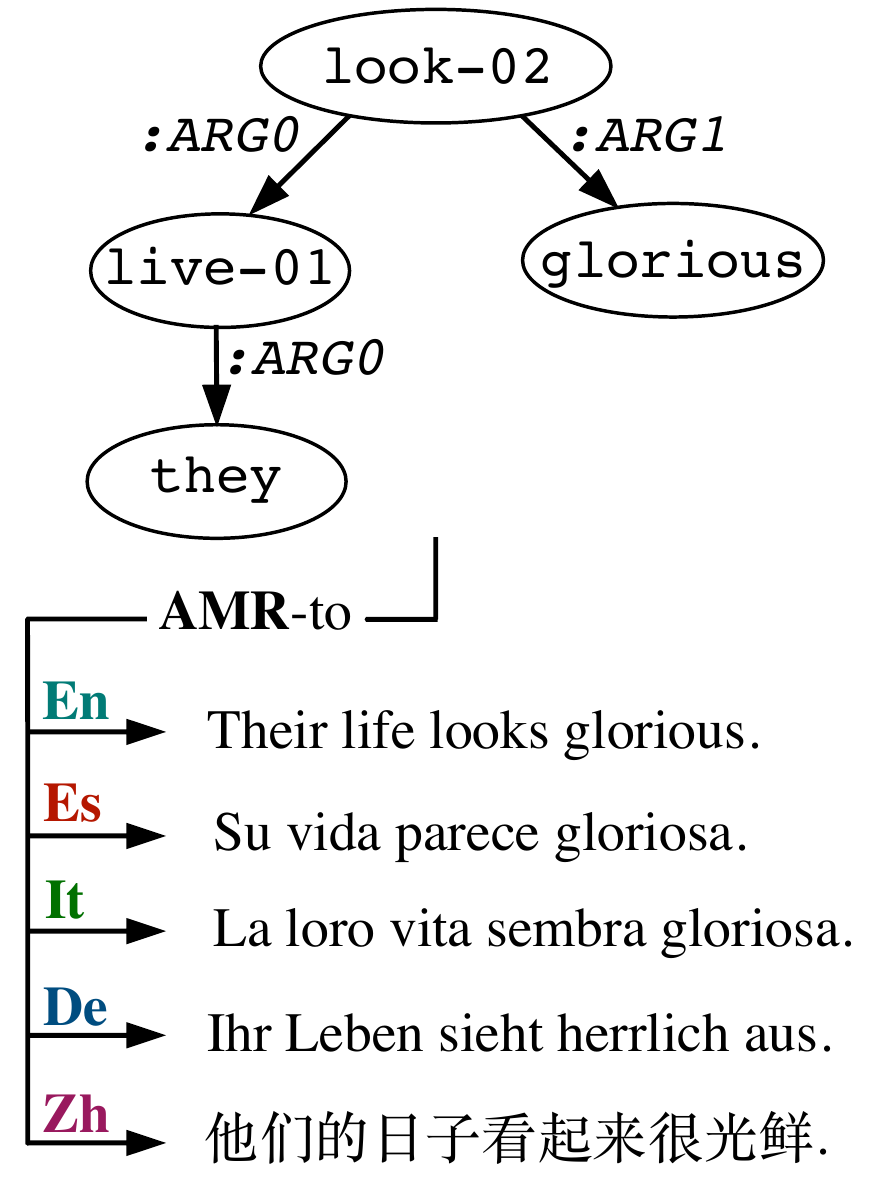}
    %\vspace{-0.5em}
    \caption{A generation example from English AMR to multiple different languages.}
    \label{fig:amrrep}
    %\vspace{-1em}
\end{figure}

While previous work has predominantly focused on monolingual English settings~\cite{,cai-lam-2020-graph, Micheleamr}, recent work has also studied  multilinguality in meaning representations \cite{blloshmi-etal-2020-xl,sheth-etal-2021-bootstrapping}.
Whereas \citet{damonte-cohen-2018-cross} demonstrate that parsers can be effectively trained to transform multilingual text into English AMR, \citet{mille-etal-2018-first,mille-etal-2019-second} and \citet{fan-gardent-2020-multilingual} discuss the reverse task, turning meaning representations into multilingual text, as shown in Figure~\ref{fig:amrrep}. However, gold-standard multilingual AMR training data is currently scarce, and previous work~\cite{fan-gardent-2020-multilingual} while discussing the feasibility of multilingual AMR-to-text generation, has investigated synthetically generated AMR as the \emph{only source} of silver training data.

In this paper, we aim to close this gap by providing an extensive analysis of different augmentation techniques to cheaply acquire silver-standard multilingual AMR-to-text data: (1) Following \citet{fan-gardent-2020-multilingual}, we parse English sentences into silver AMRs from parallel multilingual corpora (\silveramr), resulting in a dataset consisting of grammatically correct sentences with noisy AMR structures. (2) We leverage machine translation (MT) and translate the English sentences from the gold AMR-to-text corpus to the respective target languages (\silversent), resulting in a dataset with correct AMR structures but potentially unfaithful or non-grammatical sentences. (3) We experiment with utilizing the AMR-to-text corpus with both gold English AMR and sentences in multi-source scenarios to enhance multilingual training.

Our contributions and the organization of this paper are the following: First, we formalize the multilingual AMR-to-text generation setting and present various cheap and efficient alternatives for collecting multilingual training data. Second, we show that our proposed training strategies greatly advance the state of the art finding that \silversent considerably outperforms \silveramr. Third, we show that \silveramr has better relative performance in relatively larger sentences, whereas \silversent performs better for relatively larger graphs. Overall, we find that a combination of both strategies further improves the performance, showing that they are complementary for this task.

\section{Related Work}

Approaches for AMR-to-text generation predominantly  focus on English, and typically employ an encoder-decoder architecture, employing a linearized representation of the graph~\cite{konstas-etal-2017-neural, ribeiro2020investigating}. Recently, models based on the graph-to-text paradigm ~\cite{ribeiro-etal-2020-modeling,schmitt-etal-2021-modeling} improve over linearized approaches, explicitly encoding the AMR structure with a graph encoder~\cite{song-etal-2018-graph,beck-etal-2018-graph,ribeiro-etal-2019-enhancing,guo-etal-2019-densely, cai-lam-2020-graph, ribeiro2021structural}.

Advances in multilingual AMR parsing have focused on a variety of different languages such as Brazilian Portuguese, Chinese, Czech and Spanish ~\cite{hajic-etal-2014-comparing, xue-etal-2014-interlingua, migueles-abraira-etal-2018-annotating, sobrevilla-cabezudo-pardo-2019-towards}. In contrast,  little work has focused on the reverse AMR-to-text setting \cite{fan-gardent-2020-multilingual}. We aim to close this gap by experimenting with different data augmentation methods for efficient multilingual AMR-to-text generation.   

\section{Multilingual AMR-to-Text Generation}
\label{sec:multigen}
In AMR-to-text generation, we transduce an AMR graph $\mathcal{G}$ to a surface realization as a sequence of tokens $y = \langle y_1,\dots,y_{|y|} \rangle$. As input we use an English-centric AMR graph where the output $y$ can be realized in different languages (see Figure~\ref{fig:amrrep}).

 \begin{table*}[t]
\centering
\small
{\renewcommand{\arraystretch}{0.7}

\begin{tabular}{lcccccccccc}
\toprule
& \multicolumn{5}{c}{\textbf{BLEU}} & \multicolumn{5}{c}{\textbf{BERTscore}}  \\
 \midrule
 & \textsc{es} & \textsc{it} & \textsc{de} & \textsc{zh} & All & \textsc{es} & \textsc{it} & \textsc{de} & \textsc{zh} & All\\
 \midrule

% \midrule
  \multicolumn{1}{l}{MT \cite{fan-gardent-2020-multilingual}}& 21.6 & 19.6 & 15.7 & - & - & -  & - & - & - & -\\
 \multicolumn{1}{l}{Multilingual model \cite{fan-gardent-2020-multilingual}}& 21.7 & 19.8 & 15.3 & - & -  & - & - & - & - & -\\

 \midrule
 $\text{MT}$ & 27.6 & 24.2 & 19.4 & 23.3 &  23.6 &87.1 & 85.7 & 83.5 & 79.9 & 84.0 \\
 \silveramr & 23.3 & 21.2 & 16.9 & 20.1 & 20.4 & 84.5 & 83.7 & 82.0 & 76.3 & 81.6 \\
  \silversent & 28.3 & 24.3 & 18.9 & 22.2 & 23.4 & 87.3 & 85.7 & 83.5 & 79.6 & 84.0 \\
\silveramr + \goldamr & 28.2 & 24.9 & 19.4 & 22.9 & 23.9 & 87.6 & 85.9 & 83.9 & 79.5 & 84.2\\

  \silversent + \goldamr & 28.5 & 24.6 & 19.2 & 22.3 & 23.7 & 87.3 & 85.8 & 83.6 & 79.6 & 84.0\\
  \silveramr + \silversent & \textbf{30.7} & \textbf{26.4} &  \textbf{20.6} & \textbf{24.2} & \textbf{25.5} & 87.8  & \textbf{86.3} & \textbf{84.1} & \textbf{80.5} & \textbf{84.7} \\
\silveramr + \silversent + \goldamr & 30.4 & 26.1 & 20.5 & 23.4 & 25.1 & \textbf{88.0} & \textbf{86.3} & \textbf{84.1} & 80.1 & 84.6\\

\bottomrule
\end{tabular}}
\caption{Results on the multilingual LDC2020T07 test set. When training on multiple seeds, the standard deviation is between 0.1 an 0.3 BLEU. The results of our models compared to the MT baseline are statistically significant.}
\label{tab:testsetresults-ldc2020}
\vspace{-4mm}
\end{table*}

\subsection{Approach}
\label{section:approach}

We employ mT5 \cite{xue2020mt5}, a Transformer-based  encoder-decoder architecture \cite{NIPS2017_7181}, motivated by prior work~\cite{ribeiro2020investigating, ribeiro2021structural}  that leverages T5~\cite{2019t5} for AMR-to-text generation.

We define $x = \textsc{LIN}(\mathcal{G})$, where $\textsc{LIN}$ is a function that linearizes $\mathcal{G}$ into a sequence of node and edge labels using depth-first traversal of the graph~\cite{konstas-etal-2017-neural}. $x$ is encoded, conditioned on which the decoder predicts $y$ autoregressively. 

Consequently, the encoder is required to learn language agnostic representations amenable to be used in a multilingual setup for the English AMR graph; the decoder attends over the encoded AMR and  is required to generate text in different languages with varied word order and morphology. 

To differentiate between  languages, we prepend a prefix ``{\small\texttt{translate AMR to <tgt\_language>:}}'' to the AMR graph representation.\footnote{For example, for AMR-to-Spanish we use the prefix ``{\small\texttt{translate AMR to Spanish:}}''.} We add the edge labels which are present in the AMR graphs of the LDC2017T10 training set to the encoder's vocabulary in order to avoid considerable subtoken splitting -- this allows us to encode the AMR with a compact sequence of tokens and also learn explicit representations for the AMR edge labels. Finally, this multilingual approach allows us to have more AMR data on the encoder side when increasing the number of considered languages. This could be particularly helpful when using languages with little training data. 

\subsection{Data}
\label{section:data}

Since gold-standard {\it training} data for multilingual AMR-to-text generation does not exist, data augmentation methods are necessary. Given a set of gold AMR training data for English and parallel corpora between English and target languages, we thus aim to identify the best augmentations strategies to achieve multilingual generation.

As our monolingual AMR-to-text training dataset, we consider the LDC2017T10 dataset (\goldamr), containing English AMR graphs and sentences. We evaluate our different approaches on the multilingual LDC2020T07 test set by~\citet{damonte-cohen-2018-cross} consisting of gold annotations for Spanish (\textsc{es}), Italian (\textsc{it}), German (\textsc{de}) and Chinese (\textsc{zh}).\footnote{This dataset was constructed by professional translators based on  the LDC2017T10 test set.} For our multilingual parallel sentence corpus we consider data from different sources. For \textsc{es}, \textsc{it} and \textsc{de}, we use: \textbf{Europarl-v7}~\cite{koehn2005epc}, an aligned corpus of European Union parliamentary debates; \textbf{Tatoeba},\footnote{https://tatoeba.org/} a large database of example sentences and translations; and \textbf{TED2020},\footnote{https://github.com/UKPLab/sentence-transformers/tree/master/docs/datasets} a dataset of translated subtitles of TED talks. For \textsc{zh}, we use the \textbf{UM-Corpus}~\cite{conf/lrec/TianWCQOY14}.

\subsection{Creating Silver Training Data}
\label{section:silvettraindata}
We experiment with two augmentation techniques that generate silver-standard multilingual training data, described in what follows.

\vspace{1mm}
\noindent\textbf{\silveramr.} We follow~\citet{fan-gardent-2020-multilingual} and leverage the  multilingual parallel corpora described in \S \ref{section:data} and generate AMRs for the respective English sentences.\footnote{The English sentences of the parallel corpus are parsed using a state-of-the-art AMR parser~\cite{cai-lam-2020-amr}.} While the multilingual sentences are of gold standard, the AMR graphs are of silver quality. Similar to~\citet{fan-gardent-2020-multilingual}, for each target language we extract a parallel dataset of 1.9M sentences.

\vspace{1mm}
\noindent\textbf{\silversent.} We fine-tune mT5 as a translation model for English to the respective target languages, using the same parallel sentences used in \silveramr. Then, we translate the English sentences of \goldamr into the respective target languages, resulting in a multilingual dataset that consists of gold AMRs and silver sentences. The multilingual training dataset contains 36,521 examples for each target language.

\section{Experiments}

We implement our models using mT5\textsubscript{base} from HuggingFace \citep{wolf2019huggingfaces}. We use the Adafactor optimizer~\cite{pmlr-v80-shazeer18a} and employ a linearly decreasing learning rate schedule without warm-up. The hyperparameters we tune include the batch size, number of epochs and learning rate.\footnote{Hyperparameter details are in the appendix~\ref{appe:hyperparameters}.} The models are evaluated in the multilingual LDC2020T07 test set, using BLEU \cite{Papineni:2002:BMA:1073083.1073135}, METEOR \cite{Denkowski14meteoruniversal}, chrF++ \cite{popovic-2015-chrf} and BERTscore~\cite{bert-score} metrics. We compare with a MT baseline -- we generate the test set with an AMR-to-English model trained with T5~\cite{ribeiro2021structural} and translate the generated English sentences to the target language using MT. For a fair comparison, our MT model is based on mT5 and trained with the same data as the other approaches.

\paragraph{Training Strategies.}
We propose different training strategies under the setting of \S\ref{section:data} in order to investigate which combination leads to stronger multilingual AMR-to-text generation. Besides training models using \silveramr or \silversent, we investigate different combinations of multi-source training also using \goldamr.

\vspace{2mm}
\noindent\textbf{Main Results.}
Table~\ref{tab:testsetresults-ldc2020} shows our main results.\footnote{METEOR and chrF++ results can be found in Appendix Table~\ref{tab:testsetresults-ldc2020-othermetrics}.} First, \silveramr substantially outperforms \citet{fan-gardent-2020-multilingual} despite being trained on the same amount of silver AMR data. We believe this is because we utilize mT5, whereas \citet{fan-gardent-2020-multilingual} use XLM~\cite{conneau-etal-2020-unsupervised}, and our parallel data may contain different domain data. 

\silversent considerably outperforms \silveramr in all metrics, despite \silveramr consisting of two orders of magnitude more data. We believe the reasons are twofold:
Firstly, the correct semantic structure of gold AMR annotations is necessary to learn a faithful realization;
Secondly, \silversent provides examples of the same domain as the evaluation test set. 
We observe similar performance to \silversent when training on both \goldamr and \silveramr, indicating that the combination of target domain data and gold AMR graphs are necessary for downstream task performance. 
 However, training on both \goldamr and \silversent yields small gains, indicating that the respective information is adequately encoded within the silver standard dataset. 

We observe similar patterns when combining the silver standard datasets. While {\small\textsc{SilverAMR+SilverSent}} complement each other, resulting in the overall best performance, adding \goldamr does not yield any notably gains. 
These results demonstrate that \emph{both} gold AMR structure and gold sentence information are important for training multilingual AMR-to-text models, while \silversent are seemingly more important.

\begin{figure}[t]
    \centering
    \includegraphics[width=0.43\textwidth]{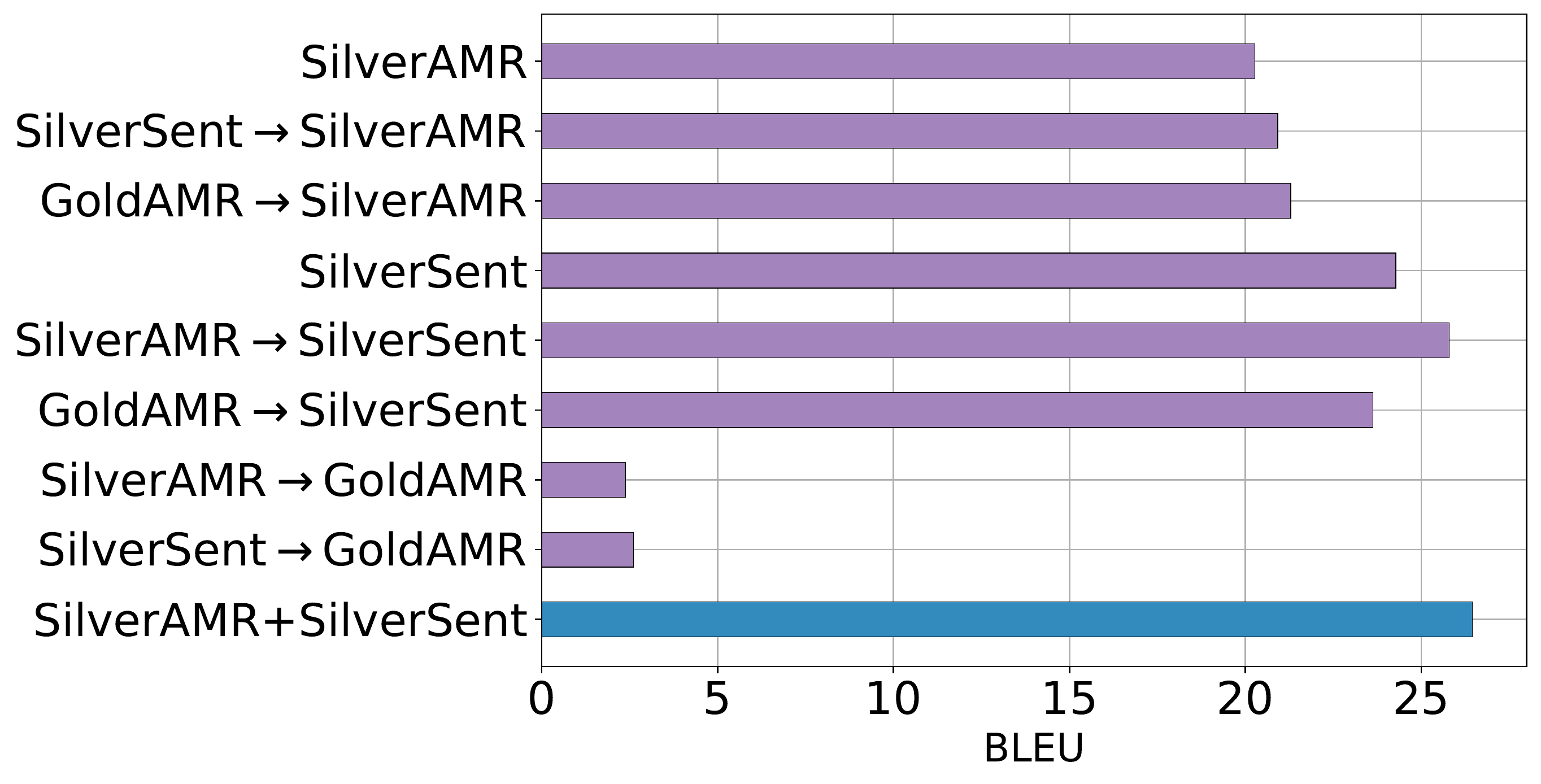}
    \caption{Order impact of sequential fine-tuning for \textsc{it}.}
    \label{fig:training}
    \vspace{-3mm}
\end{figure}

\vspace{2mm}
\noindent\textbf{Effect of the Fine-tuning Order.} In  Figure~\ref{fig:training} we illustrate the impact of different data source orderings when fine-tuning in a two-phase setup for \textsc{it}.\footnote{Other languages follow similar trends and are presented in Figure~\ref{fig:alltraining} in the Appendix.} Firstly, we observe a decrease in performance for all sequential fine-tuning settings, compared to our proposed mixed multi-source training, which is likely due to \emph{catastrophic forgetting}.\footnote{The model trained on the second task forgets the first task.} Secondly, training on \silveramr and subsequently on \silversent (or vice versa), improves performance over only using either, again demonstrating their complementarity. Thirdly, \silversent continues to outperform \silveramr as a second task. Finally, \goldamr is not suitable as the second task for multilingual settings as the model predominantly generates English text.

\begin{figure}[t]
    \centering
    \includegraphics[width=0.34\textwidth]{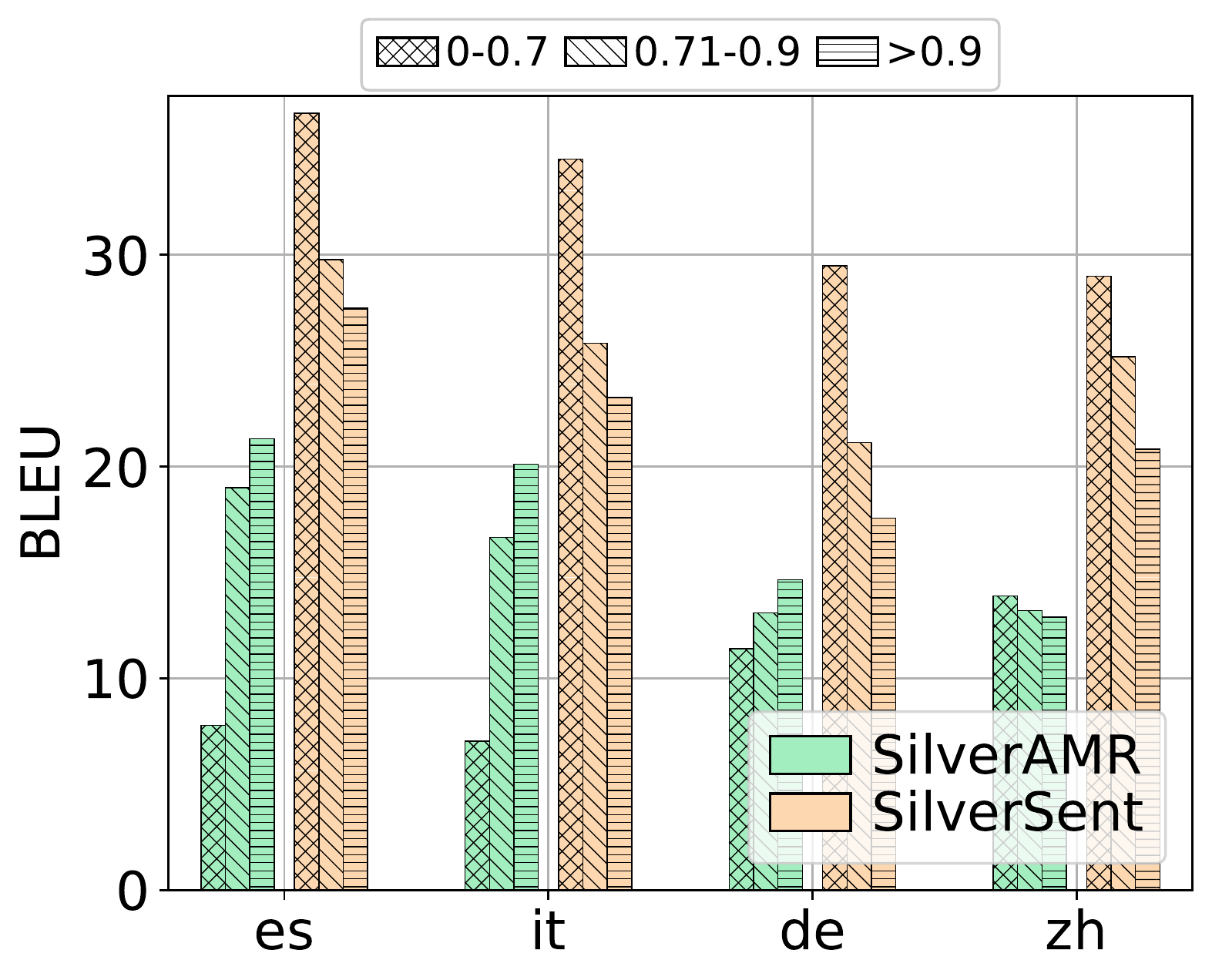}
    \caption{Impact of the sentence length and graph size ratio $\gamma$ on the LDC2020T07 multilingual test set.}
    \label{fig:ratio}
\end{figure}

\vspace{2mm}
\noindent\textbf{Impact of Sentence Length and Graph Size.} 
As silver annotations potentially lead to noisy inputs, models trained on \silveramr are potentially less capable of encoding the AMR semantics correctly, and models trained on \silversent potentially generate fluent sentences less reliably. 
To analyze the advantages of the two forms of data, we measure the performance against the sentence lengths and graph sizes.\footnote{Sentence lengths were measured using subwords.} We define $\gamma$ to be a ratio of the  
sentence length, divided by the number of AMR graph nodes. In Figure~\ref{fig:ratio} we plot the respective results for \silveramr and \silversent, categorized into three bins. We find that almost all \silveramr's BLEU increases for longer sentences, suggesting that training with longer gold \textit{sentences} improves performance. In contrast, with larger \textit{graphs}, the BLEU performance improves for \silversent, indicating that large gold AMR graphs are also important. \silveramr and \silversent present relative gains in performance on opposite ratios of sentence length and graph size, suggesting that they capture distinct aspects of the data. 

\begin{table}[t]
\small
\centering
{\renewcommand{\arraystretch}{0.6}

\begin{tabular}{l@{\hspace*{3mm}}c@{\hspace*{2.6mm}}c@{\hspace*{2.6mm}}c@{\hspace*{2.6mm}}c} 
\toprule
 & \textsc{es} & \textsc{it} & \textsc{de} & \textsc{zh}  \\
\midrule
\silveramr & 19.3 & 16.5 & 11.8 & 11.9 \\
\silversent & 22.3 & 17.3 & 12.7 & 11.9\\
\silveramr + \silversent & 23.5 & 19.2 & 15.0 & 13.0 \\
\bottomrule
\end{tabular}}
\caption{BLEU results for out of domain evaluation.}
\label{tab:outdomain}
\vspace{-3mm}
\end{table}

\begin{table*}[t]
\small
\begin{center}
\begin{tabular}{lp{11cm}}
\toprule
 \textbf{Model}       & \textbf{Examples}    \\
\midrule
AMR &  (m / multi-sentence \\
   &  \quad\quad :snt1 (w2 / wish-01  \\
&     \quad\quad\quad\quad       :ARG0 (i2 / i)  \\
&   \quad\quad\quad\quad         :ARG1 (p / possible-01  \\
&   \quad\quad\quad\quad\quad\quad               :ARG1 (w3 / wipe-out-02 \\
& \quad\quad\quad\quad\quad\quad\quad\quad                        :ARG0 i2 \\
&  \quad\quad\quad\quad\quad\quad\quad\quad                      :ARG1 (s / she) \\
&  \quad\quad\quad\quad\quad\quad\quad\quad                       :source (l / live-01 \\
&  \quad\quad\quad\quad\quad\quad\quad\quad\quad\quad                           :ARG0 i2)))) \\
&  \quad\quad    :snt2 (g / good-02 \\
&  \quad\quad \quad\quad          :ARG1 (t / thing) \\
&  \quad\quad \quad\quad          :degree (m2 / more \\
&  \quad\quad \quad\quad \quad\quad                :degree (m3 / much \\
&  \quad\quad \quad\quad \quad\quad \quad\quad                      :degree (s2 / so))) \\
&  \quad\quad \quad\quad          :prep-without (s3 / she))) \\[0.3cm]
 \silveramr  & \textbf{\textcolor{red}{Con ella, las cosas son mucho mejor}}. Deseo que pudiera eliminarla de mi vida.\\[0.3cm]
 \silversent & Desearía \textbf{{\color{red}que podía}} eliminarla de mi vida. Las cosas serían mucho mejor sin ella. \\[0.3cm]
 {\small\textsc{SilverAMR+SilverSent}} &  Desearía poder eliminarla de mi vida, las cosas serían mucho mejor sin ella.\\[0.3cm]
 Reference & Ojalá pudiera borrarla de mi vida, las cosas hubieran sido mucho mejor sin ella.\\[0.3cm]
 English Reference & I wish I could wipe her out of my life - things would be so much better without her.
\\

\midrule
\end{tabular} 
\caption{Example of an AMR, generated texts in \textsc{es} by the different models, and its \textsc{es} and \textsc{en} references. We indicate in \textbf{{\color{red}red}} errors (unfaithfulness in \silveramr and incorrect grammar in \silversent) that are not present in {\small\textsc{SilverAMR+SilverSent}} and in the human-written reference.}
\label{tab:generatedexamples}
\end{center}
\end{table*} 

\vspace{2mm}
\noindent\textbf{Out of Domain Evaluation.} 
To disentangle the effects of in-domain sentences and gold quality AMR graphs in \silversent,
we evaluate both silver data approaches on the \textbf{Weblog and WSJ} subset of the LDC2020T07 dataset; The domain of this subset is \emph{not included} in the LDC2017T10 training set. 
We present the BLEU results in Table~\ref{tab:outdomain}.\footnote{BERTscore results can be found in Appendix Table~\ref{tab:outdomain-bert}.}  
While we find that \silversent prevails in achieving better performance --- demonstrating that AMR gold structures are an important source for training multilingual AMR-to-text models --- \silveramr and \silversent perform more comparably than when evaluated on the full LDC2020T07 test set. This
demonstrates that the domain transfer factor plays an important role in the strong performance of \silversent.  Overall, {\small\textsc{SilverAMR+SilverSent}} outperforms both single source settings, establishing  the complementarity of both silver sources of data.

\vspace{2mm}
\noindent\textbf{Case Study.}
Table~\ref{tab:generatedexamples} shows an AMR, its reference sentences in \textsc{es} and \textsc{en}, and sentences generated in \textsc{es} by \silveramr, \silversent, and their combination. The incorrect verb tense is due to the lack of tense information in AMR. \silveramr fails in capturing the correct concept \emph{prep-without} generating an unfaithful first sentence. This demonstrates a potential issue with approaches trained with silver AMR data where the input graph structure can be noisy, leading to a model less capable of encoding AMR semantics. On the other hand, \silversent correctly generates sentences that describe the graph, while it still generates a grammatically incorrect sentence (wrongly generating \emph{que podía} after \emph{desearía}). This highlights a potential problem with approaches that employ silver sentence data where sentences used for the training could be ungrammatical, leading to models less capable of generating a fluent sentence. Finally, {\small\textsc{SilverAMR+SilverSent}} produces a more accurate output than both silver approaches by generating grammatically correct and fluent sentences, correct pronouns, and mentions when control verbs and reentrancies (nodes with more than one entering edge) are involved.

\section{Conclusion}
The unavailability of gold training data makes multilingual AMR-to-text generation a challenging topic. We have extensively evaluated data augmentation methods by leveraging existing resources, namely a set of gold English AMR-to-text data and a corpus of multilingual parallel sentences. Our experiments have empirically validated that both sources of silver data --- silver AMR with gold sentences and gold AMR with silver sentences --- are complementary, and a combination of both leads to state-of-the-art performance on multilingual AMR-to-text generation tasks.

\section*{Acknowledgments}

We would like to thank G{\"o}zde G{\"u}l Sahin, Ji-Ung Lee, Kevin Stowe, Kexin Wang and Nandan Thakur for their feedback on this work.  Leonardo F. R. Ribeiro is supported by the German Research Foundation (DFG) as part of the Research Training Group ``Adaptive Preparation of Information form Heterogeneous Sources'' (AIPHES, GRK 1994/1) and as part of the DFG funded project UKP-SQuARE with the number GU 798/29-1. Jonas Pfeiffer is supported by the LOEWE initiative (Hesse, Germany) within the emergenCITY center.

\bibliography{anthology,custom}
\bibliographystyle{acl_natbib}

%\clearpage
\appendix

\section*{Appendices}

\section{Details of Models and Hyperparameters} % (fold)
\label{appe:hyperparameters}
The experiments were executed using the version $4.4.0$ of the \emph{transformers} library by Hugging Face \citep{wolf2019huggingfaces}. Table \ref{tab:hyper} shows the hyperparameters used to train our models. BLEU is used for model selection using translated sentences of the LDC2017T10 development set. We train until the results on the development set BLEU have not improved for 6 epochs.

\begin{table}[h]
\centering
\begin{tabular}{lc}
\toprule
         learning rate & 1e-04 \\
         batch size & 8\\
         beam search size & 6  \\
         max source length & 350  \\
         max target length & 200  \\

\bottomrule
\end{tabular}
\caption{Hyperparameter settings for our methods. }
\label{tab:hyper}
\end{table}

\section{Main Results: Additional Metrics}

In Table~\ref{tab:testsetresults-ldc2020-othermetrics} we present additional results on the multilingual LDC2020T07 test set using METEOR \cite{Denkowski14meteoruniversal}, chrF++ \cite{popovic-2015-chrf} metrics.

\section{Results: Out of Domain Evaluation}

In Table~\ref{tab:outdomain-bert} we show BERTscore~\cite{bert-score} results for out of domain evaluation on the \textbf{Weblog and WSJ} subset of the LDC2020T07 dataset.

\begin{table}[h]
\small
\centering
{\renewcommand{\arraystretch}{0.6}

\begin{tabular}{l@{\hspace*{3mm}}c@{\hspace*{2.6mm}}c@{\hspace*{2.6mm}}c@{\hspace*{2.6mm}}c} 
\toprule
 & \textsc{es} & \textsc{it} & \textsc{de} & \textsc{zh}  \\
\midrule
\silveramr & 83.3 & 81.2 & 79.8 & 73.6 \\
\silversent & 84.6 & 83.0 & 80.4 & 73.0\\
\silveramr + \silversent & 84.6 & 83.2 & 81.2 & 74.1 \\
%Adapter  & 39.05 & & \\
%AdapterGNN  & 40.71  & & \\
\bottomrule
\end{tabular}}
\caption{BERT scores for out of domain evaluation.}
\label{tab:outdomain-bert}
\vspace{-3mm}
\end{table}

\section{Results: Sequential Fine-tuning}

\begin{figure*}[t]
    \centering
    \includegraphics[width=\textwidth]{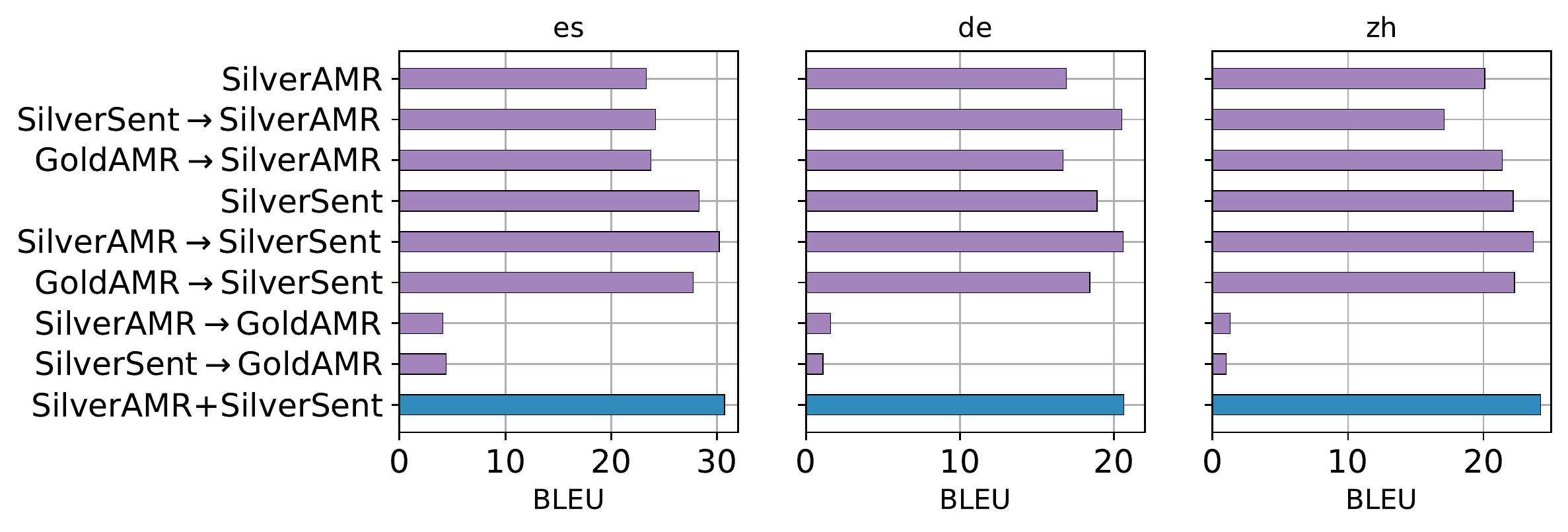}
    \caption{Order impact of sequential fine-tuning in the LDC2020T07 test set for \textsc{es}, \textsc{de} and \textsc{zh}.}
    \label{fig:alltraining}
\end{figure*}

In Figure~\ref{fig:alltraining} we present the impact of sequential fine-tuning strategies in the LDC2020T07 test set for \textsc{es}, \textsc{de} and \textsc{zh}.

\begin{table*}[t]
\centering
\small
{\renewcommand{\arraystretch}{0.7}

\begin{tabular}{lcccccccccc}
\toprule
& \multicolumn{5}{c}{\textbf{METEOR}} & \multicolumn{5}{c}{\textbf{chrF++}}  \\
 \midrule
 & \textsc{es} & \textsc{it} & \textsc{de} & \textsc{zh} & All & \textsc{es} & \textsc{it} & \textsc{de} & \textsc{zh} & All\\
 \midrule
 $\text{MT}$ & 29.9 & 27.2 & 23.2 & 25.7 & 26.5 & 54.8 & 52.0 & 47.3 & 22.3 & 44.1\\
 \silveramr & 28.3 & 26.0 & 22.7 & 23.3 & 25.0 & 51.3 & 49.6 & 45.9 & 19.5 & 41.5\\
  \silversent & 30.6 & 27.3 & 23.0 & 24.9 & 26.4 & 55.6 & 52.2 & 47.2 & 21.7 & 44.1\\
\silveramr + \goldamr & 29.8 & 26.9 & 23.6 & 25.2 & 26.3 & 55.9 & 51.7 & 47.5 & 22.3 & 44.3\\
  \silversent + \goldamr & 30.4 & 27.5 & 23.3 & 24.9 & 26.5 & 55.3 & 52.3 & 47.3 & 21.8 & 44.1\\
  \silveramr + \silversent & \textbf{31.9} & \textbf{28.7} & \textbf{24.4} & \textbf{26.4} & \textbf{27.8} & \textbf{57.2} & \textbf{54.0} & \textbf{49.4} & \textbf{23.0} & \textbf{45.9}\\

\silveramr + \silversent + \goldamr & 31.7 & 28.6 & 24.2 & 25.7 & 27.5 & \textbf{57.2} & 53.6 & 48.6 & 22.5 & 45.4 \\

\bottomrule
\end{tabular}}
\caption{METEOR and chrF++ results on the multilingual LDC2020T07 test set.}
\label{tab:testsetresults-ldc2020-othermetrics}

\end{table*}

\end{document}